\documentclass[letterpaper]{article} 
\usepackage{aaai25}  
\usepackage{times}  
\usepackage{helvet}  
\usepackage{courier}  
\usepackage[hyphens]{url}  
\usepackage{graphicx} 
\usepackage{natbib}  
\usepackage{caption} 
\usepackage{algorithm}
\usepackage{algorithmic}

\usepackage{newfloat}
\usepackage{listings}
\DeclareCaptionStyle{ruled}{labelfont=normalfont,labelsep=colon,strut=off} 
\floatstyle{ruled}
\newfloat{listing}{tb}{lst}{}
\floatname{listing}{Listing}
\usepackage{tcolorbox}
\usepackage{amsmath}
\usepackage{booktabs}
\title{Augmenting Math Word Problems via Iterative Question Composing}
\author{
Haoxiong Liu\textsuperscript{\rm 1}\equalcontrib\thanks{Corresponding author.},
Yifan Zhang\textsuperscript{\rm 1}\equalcontrib,
Yifan Luo\textsuperscript{\rm 1 \rm 2},
Andrew Chi-Chih Yao\textsuperscript{\rm 1 \rm 2}
}

\affiliations{
\textsuperscript{\rm 1}Institute for Interdisciplinary Information Sciences, Tsinghua University, \textsuperscript{\rm 2}Shanghai Qizhi Institute \\
    \{liuhx20,zhangyif21,luoyf24\}@mails.tsinghua.edu.cn, andrewcyao@tsinghua.edu.cn
}

\usepackage{bibentry}
\begin{document}

\maketitle

\begin{abstract}
Despite the advancements in large language models (LLMs) for mathematical reasoning, solving competition-level math problems remains a significant challenge, especially for open-source LLMs without external tools. We introduce the MMIQC dataset, comprising a mixture of processed web data and synthetic question-response pairs, aimed at enhancing the mathematical reasoning capabilities of base language models. Models fine-tuned on MMIQC consistently surpass their counterparts in performance on the MATH benchmark across various model sizes. Notably, Qwen-72B-MMIQC achieves a 45.0\% accuracy, exceeding the previous open-source state-of-the-art by 8.2\% and outperforming the initial version GPT-4 released in 2023. Extensive evaluation results on Hungarian high school finals suggest that such improvement can generalize to unseen data. Our ablation study on MMIQC reveals that a large part of the improvement can be attributed to our novel augmentation method, Iterative Question Composing (IQC), which involves iteratively composing new questions from seed problems using an LLM and applying rejection sampling through another LLM.
\end{abstract}

%
\begin{links}
    \link{Code}{https://github.com/iiis-ai/IterativeQuestionComposing}
    \link{Datasets}{https://huggingface.co/datasets/Vivacem/MMIQC}
\end{links}

\begin{figure}[ht]
\begin{center}
\centerline{\includegraphics[width=0.5\textwidth]{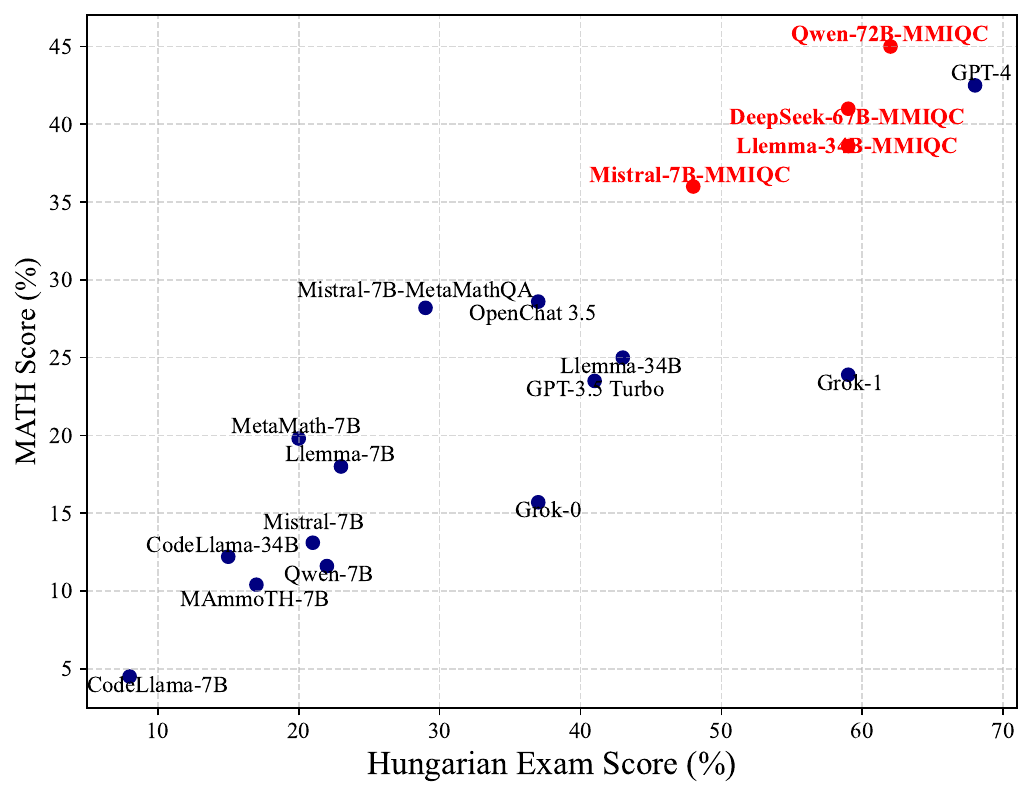}}
\caption{Performance evaluation of various LLMs on MATH~\cite{hendrycks2021measuring} and the 2023 Hungarian National High School Mathematics Finals~\cite{testing_language_models_on_a_held_out_high_school_national_finals_exam}.}
\label{fig:hungarian}
\end{center}
\end{figure}

\begin{figure*}[ht]
\begin{center}
\centerline{\includegraphics[width=1.0\textwidth]{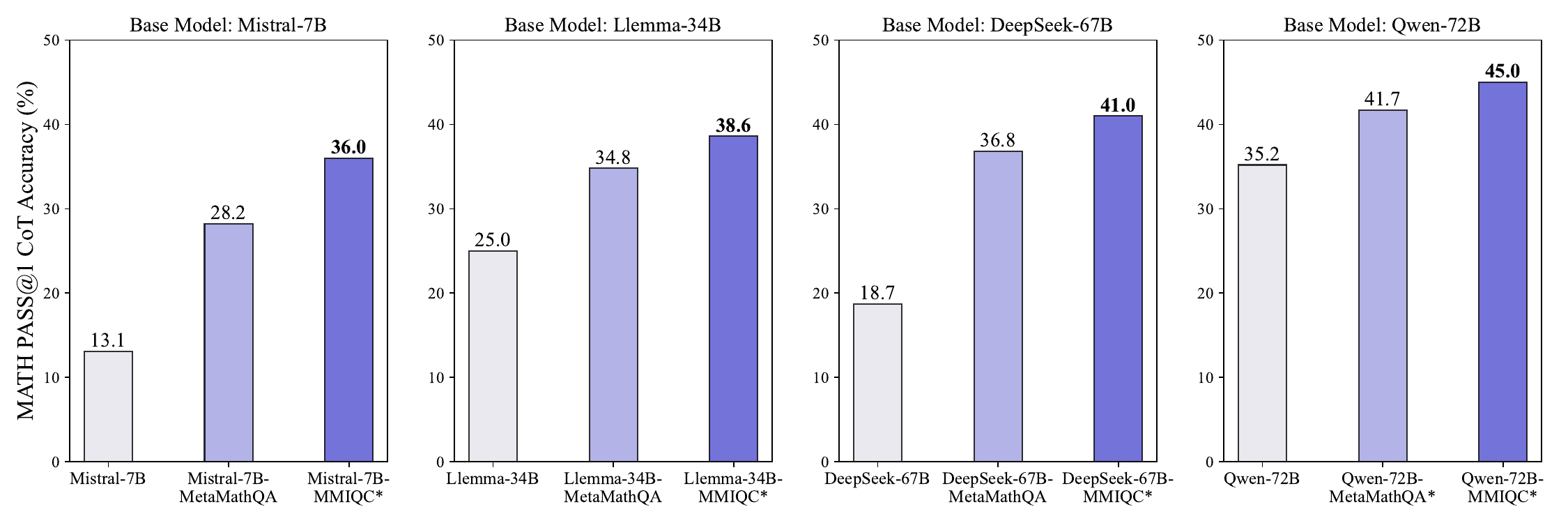}}
\caption{The performance of base models and their fine-tuned versions on MATH benchmark. The models remarked with an $^*$ are trained and evaluated by us. We can see that the models fine-tuned on MMIQC consistently outperform their counterparts by a clear margin.}
\label{fig:math-acc}
\end{center}
\end{figure*}

\section{Introduction}
\label{sec:intro}
Although large language models have been demonstrated to be powerful in various applications~\cite{chen2021evaluating, brown2020gpt3, ouyang2022training, park2023generative, Huang2022LanguageMA}, solving math problems that require complex reasoning skills remains a challenging task. On MATH~\cite{hendrycksmath2021}, a competition-level math problem benchmark containing algebra, calculus, geometry, combinatorics and number theory problems, open-source base LLMs such as the LLaMA family~\cite{touvron2023llama, Touvron2023Llama2O} fail to answer most of the problems correctly. 

Previous work tries to enhance the mathematical reasoning abilities of base models by fine-tuning them on domain-specific data. Specifically, One line of work ~\cite{azerbayev2023llemma, lewkowycz2022solving} collects math corpora from the web and fine-tunes the models on them, which is also known as the procedure of continual pre-training~\cite{cossu2022continual}. Another line of work focuses on constructing synthetic data through rejection sampling ~\cite{yuan2023scaling}, distilling from GPT-4/GPT-3.5~\cite{yue2023mammoth} or question bootstrapping~\cite{yu2023metamath}, and then use the generated question-response pairs to perform supervised fine-tuning in the way described in~\cite{alpaca, ouyang2022training}. However, there still exists a large performance gap between these fine-tuned models and the most advanced close-source models such as GPT-4~\cite{OpenAI2023GPT4TR} and Gemini-Ultra~\cite{team2023gemini}. Given that simply adding more data does not always lead to better performance as shown in ~\cite{yu2023metamath}, how to bridge the gap remains an open challenge.

This work tackles the challenge by combining the two lines of work. On one hand, we reuse the high-quality corpora used in the pre-training stage during fine-tuning. Specifically, MMIQC contains around 1200k question-response pairs we filtered and pre-processed from the web pages at math.stackexchange.com, which are included in the RedPajama dataset~\cite{together2023redpajama}. On the other hand, for the synthetic data part of MMIQC, we increase the diversity by using multiple kinds of augmentation methods listed below: 1)~Prompting GPT-4 with an integrated version of the question bootstrapping prompts used in \cite{yu2023metamath}, and do rejection sampling with GPT-3.5-Turbo on both seed and augmented problems. 2)~Using a modified prompt presented in \cite{liu2023tinygsm} to ask GPT-4 to generate similar problems with answers given seed problems of the training set of MATH. Although the generated answers can be wrong, we perform rejection sampling on these problems as well. 3)~Performing IQC (Iterative Question Composing) with 4 iterations in total. We iteratively ask GPT-4 to compose new questions from the given seed problems and do rejection sampling to filter those problems with answers aligned with GPT-3.5-turbo's answers. 4)~Filtering a 204k subset of MetaMathQA~\cite{yu2023metamath} and adding it to the MMIQC dataset (More details on MMIQC will be introduced in Section~\ref{subsec:mmiqc}).

We fine-tune several base models on MMIQC, resulting in models consistently achieving a large margin compared to their counterparts when evaluated on MATH, as shown in Figure~\ref{fig:math-acc}.
Specifically, the models Mistral-7B-MMIQC, Llemma-34B-MMIQC, DeepSeek-67B-MMIQC and Qwen-72B-MMIQC, which are obtained by fine-tuning Mistral-7B~\cite{jiang2023mistral}, Llemma-34B~\cite{azerbayev2023llemma} and DeepSeek-67B~\cite{bi2024deepseek} on MMIQC, achieve 36.0\%, 38.6\%, 41.0\% and 45.0\% accuracy on MATH, 5.8\%, 3.8\%, 4.2\% and 3.3\% higher than the counterpart models that are fine-tuned on MetaMathQA, respectively. 

We also evaluate the models on the 2023 Hungarian national high school finals in mathematics \cite{Keiran2024hungarian}. The results in Figure~\ref{fig:hungarian} suggest that the mathematical reasoning abilities the models acquire through being fine-tuned on MMIQC can generalize to unseen held-out problems.

We highlight our contributions as follows:
\begin{itemize}
    \item We propose IQC (Iterative Question Composing), a data augmentation method that can iteratively generate diverse data starting from a seed dataset of math word problems.
    \item We release MMIQC, a mixture of processed web data and synthetic question-response pairs. In different model sizes, the models fine-tuned on MMIQC consistently outperform their counterparts by a clear margin on the MATH test set. Notably, Qwen-72B-MMIQC achieves a 45.0\% accuracy, exceeding the previous open-source state-of-the-art\footnote{As of the time of writing in January 2024, to the best of our knowledge, the open-source SOTA on MATH is the DeepSeek-67B-MetaMathQA model reported in \cite{wang2023math}, which achieves 36.8\% accuracy without external tool usage.} by 8.2\% and outperforming the initial version GPT-4 released in 2023. Such improvement can generalize to unseen held-out data, e.g., Hungarian high school finals.
    \item Our results show that reusing the high-quality data in the pre-training corpora during the fine-tuning stage can improve the model performance, successfully combining the two lines of work of continual pre-training and supervised fine-tuning.
    \item Our results also show that using multiple augmentation methods to construct datasets for fine-tuning is an efficient way to boost the performance of LLMs. 
\end{itemize}

\section{Related Work}

\textbf{Base Large Language Models.} Base large language models (LLMs) trained on massive corpora (e.g. 1.4T tokens of text for Llama~\cite{touvron2023llama}) from various sources with a simple auto-regressive next token prediction loss have achieved great success in various natural language processing tasks~\cite{radford2019language, brown2020gpt3, touvron2023llama, Touvron2023Llama2O, jiang2023mistral}. Although these pre-trained models are not intended to serve for solving complex mathematical problems, \cite{wei2023chainofthought} show that few-shot prompting can help the models answer a certain fraction of problems correctly. Nevertheless, to achieve better performance, fine-tuning the base LLMs on domain-specific data is required.

\textbf{Fine-tuning Base LLMs on Mathematical Datasets. } Current practice of fine-tuning base LLMs on mathematical datasets can be classified into two kinds: 1) continual pretraining~\cite{lewkowycz2022solving, azerbayev2023llemma}. This line of work typically collects billion-tokens level mathematical text data from the web, such as mathematical sub-sites of Stack Exchange and ArXiv, and fine-tune the model in the same way as that in the pre-training stage. 
2) SFT (Supervised Fine-Tuning)~\cite{yuan2023scaling, yu2023metamath, yue2023mammoth, gou2023tora}. Works in this line collect question-response pairs via various methods and train the models on their dataset in an Alpaca style. Due to the scarcity of publicly available high-quality question-response pairs datasets and the costly nature of manually composing math word problems, how to augment new data from the existing datasets becomes the focus of these works.

Our work is located in the middle between these two: MMIQC is a mixture of filtered pre-training corpus and question-response pairs generated using various augmentation methods. 

\textbf{Reasoning Frameworks for Solving Mathematical Problems. } Much effort has been devoted to achieving a higher accuracy on math word problem benchmarks by designing different procedures of using the given LLMs to obtain the answers, which we refer to as \textit{reasoning frameworks}. Among them, \textit{Prompting-based} methods \cite{radford2019language, wei2023chainofthought, fu2022complexity} play a significant role in activating the potential reasoning abilities for base LLMs through carefully designing the prompts shown to the models. Self-consistency~\cite{wang2023selfconsistency} samples multiple rationale paths for a model and then decides the answer by majority voting.  
In contrast of self-consistency, \cite{cobbe2021gsm8k, uesato2022solving, lightman2023lets} use Outcome Reward Models (ORM) and Process Reward Models (PRM) trained on human annotations as verifiers to help select the answer with the highest score from the sampled reasoning paths of LLMs. Getting rid of the need of manual annotation, \cite{wang2023math} score a given reasoning step by estimating the potential of that step to lead to a correct answer automatically. 
 
Some frameworks also include the use of plug-in tools and external APIs. Program-aided prompting~\cite{gao2022pal, yue2023mammoth} provides in-context samples containing Python codes for LLMs and uses code interpreters to execute the output to facilitate reasoning. Further, \cite{gou2023tora} interleave natural language rationales with {Sympy}\footnote{https://www.sympy.org/} code and fine-tune the model on trajectories sampled from GPT-4 to follow their framework in two steps, namely imitation learning and output space shaping.

We note that our results in Figure~\ref{fig:math-acc} do not include multiple times of sampling, use of verifiers or code interpreters, thus cannot be directly compared with the results reported in these works.

\section{Iterative Question Composing}
\label{subsec:iqc}
\begin{algorithm}[tb]
   \caption{Iterative Question Composing}
   \label{alg:iqc}
\begin{algorithmic}[1]
\REQUIRE Question composing model $\pi_q$, rejection sampling model $\pi_r$, answer extractor defining $\simeq$, text templater $x(\cdot, \cdot)$ with inverse $x^{-1}(\cdot)$, initial seed dataset $S_0 = \{(q_i, a_i)\}_{i=1}^n$, total iterations $K$, question composing prompts $p_1, p_2, \ldots, p_K$, rejection sampling prompt $p_r$, maximum rejection samples per problem $m$
\FOR{$k=1$ \textbf{to} $K$}
\STATE Initialize $S_k \leftarrow \{\}$, $R_k \leftarrow \{\}$
\FORALL{$(q,a) \in S_{k-1}$}
\STATE Sample $x' \sim \pi_q\left(\cdot | p_k \oplus x(q,a)\right)$
\STATE Decompose $(q', a') \leftarrow x^{-1}(x')$
\STATE Append $S_k \leftarrow S_k \cup \{(q', a')\}$
\FOR{$j=1$ \textbf{to} $m$}
\STATE Sample $a^{(j)} \sim \pi_r(\cdot | p_r \oplus q')$
\IF{$a^{(j)} \simeq a'$}
\STATE Append $R_k \leftarrow R_k \cup \{(q', a^{(j)})\}$
\ENDIF
\ENDFOR
\ENDFOR
\STATE Combine $D_k \leftarrow S_k \cup R_k$
\ENDFOR
\STATE Output Collections $D_1, D_2, \ldots, D_K$
\end{algorithmic}
\end{algorithm}

Traditional data augmentation methods primarily concentrate on modifying either the questions or answers while retaining their original meanings, or generating similar problems, as discussed in \cite{yu2023metamath} and \cite{liu2023tinygsm}. These methods, however, are limited in their diversity as they aim to create nearly identical problems. Our approach, termed \textbf{IQC (Iterative Question Composing)}, deviates from this by iteratively constructing more complex problems. It augments the initial problems, adding additional reasoning steps without altering their intrinsic logical structure. This ensures that the newly formed problems are organically linked to the original problem and elaborately tries to not include extraneous elements induced by a large transition of the reasoning process.

\textbf{Notations. } In our description, we refer to the combination of an LLM, its tokenizer, encoding/decoding methods, and a fixed generation configuration (inclusive of generation strategy, sampling temperature, and stopping criteria) simply as `an LLM'. For an LLM $\pi$, we denote the output distribution given prompt $p \in \mathcal{A}^*$ as $\pi(\cdot | p)$. The concatenation of two text paragraphs $p_1$ and $p_2$ is represented as $p_1\oplus p_2$.

The IQC process begins with specifying an LLM $\pi_q$ for question composing and another model $\pi_r$ for rejection sampling. An answer extractor is needed to derive answers from responses. Two responses $r_1$ and $r_2$ are considered equivalent, denoted $r_1\simeq r_2$, if the same answer can be extracted from both. The process initiates with a seed dataset $S_0 = \{(q_i, a_i)\}_{i=1}^n$.

In iteration \#1, we prompt $\pi_q$ with $p_1\oplus x(q, a)$ for each $(q, a)\in S_0$, where $x(\cdot, \cdot)$ is a text template transforming a question-response pair into text, and $p_1$ solicits a new question-answer composition. This yields a new dataset
\begin{equation*}
    S_1 = \{(q'_i, a'_i)\}_{i=1}^n,
\end{equation*}
where $(q'_i, a'_i) = x^{-1}(x'_i)$ and $x'_i \sim \pi_q\left(\cdot | p_1\oplus x_i\right)$ is the output for the $i$th sample. We further enhance $S_1$ by rejection sampling from $\pi_r$, resulting in
\begin{equation*}
    R_1 := \{(q'_i, a^{(j)}_i)| a^{(j)}_i \simeq a'_i,  i\in[n], j \in [m]\},
\end{equation*}
where $a^{(j)}_i$ are the sampled responses from $\pi_r(\cdot|p_r\oplus q'_i)$. The dataset $D_1$ is then formed by uniting $S_1$ and $R_1$:
\begin{equation*}
    D_1 := S_1 \cup R_1.
\end{equation*}

For each subsequent iteration \#$k$, the aforementioned procedure is repeated using $S_{k-1}$ as the seed dataset, with varying question composing prompts $p_k$. The complete IQC process is delineated in Algorithm~\ref{alg:iqc}.

\begin{figure}[ht]
\begin{tcolorbox}
\textbf{Seed Question:}\\ Evaluate \begin{align*} (5a^2 - 13a + 4)(2a - 3) \end{align*} for $a = 1\frac12$.
\vspace{2mm}

\textbf{Iter \#~1 Question:}\\ If $b = 2a - 3$ and $a = 1\frac12$, what is the value of $(5a^2 - 13a + 4)b$?
\vspace{2mm}

\textbf{Iter \#~2 Question:}\\ Given $b = 2a - 3$, $a = 1\frac12$, and $c = 3b + 5$, find the value of $c(5a^2 - 13a + 4)$.
\vspace{2mm}

\textbf{Iter \#~3 Question:}\\ Given $b = 2a - 3$, $a = 1\frac12$, $c = 3b + 5$, and $d = c^2 - 4c$, find the value of $d + c(5a^2 - 13a + 4)$.
\vspace{2mm}

\textbf{Iter \#~4 Question:}\\ Given $b = 2a - 3$, $a = 1\frac12$, $c = 3b + 5$, $d = c^2 - 4c$, and $e = d^3 + 2cd - 7$, find the value of $e + c(5a^2 - 13a + 4) + d$.

\end{tcolorbox}
\caption{An example of the questions composed via IQC by GPT-4 given 1 seed problem in MATH training set.}
\label{fig:iqc_example}
\end{figure}

\begin{figure}[ht]
\begin{tcolorbox}
You will be provided with 1 math problem and its solution and answer \textit{(which are not guaranteed to be right)}. Please generate 1 new problem that (implicitly) contains the original problem as a subproblem or substep.\\

Your response should only contain one line text with 3 fields "problem", "solution" and "answer" in the same format as the given problem. The solution to the generated problem should be as brief as possible and **should not quote the conclusion of the original problem**. Ensure there is only one latex box in the solution and the answer is completely the same with the content in the box.\\

**Please use two backslashes to represent one in the strings in order that it can be properly read in python.** For example, you should write ``\textbackslash cdot'' as ``\textbackslash\textbackslash cdot''.
\end{tcolorbox}
\caption{The prompt we use to perform question composing in IQC. The \textit{italics} part is not used in iteration~\#1.}
\label{fig:qc_prompt}
\end{figure}

\section{The MMIQC Dataset}
\label{subsec:mmiqc}

In this section, we introduce how each part of MMIQC is constructed in detail. 

\textbf{Subset of MetaMathQA. } The original MetaMathQA dataset is constructed by sampling GPT-3.5 for $k=20$ times under a $T=0.7$ temperature for each problem in the training set of MATH~\cite{hendrycks2021measuring} and GSM8K~\cite{cobbe2021gsm8k} dataset, or its bootstrapped versions. We restrict the number of samples for each completely same question to be 3 and 1 for MATH and GSM8K, respectively, to obtain a subset of MetaMathQA. This subset contains 112.2K GSM8K question-response pairs and 91.5K MATH pairs.

\textbf{Answer Augmentation and Question Bootstrapping. } We integrate the question bootstrapping methods used in \cite{yu2023metamath} into a single prompt shown in Figure~\ref{fig:qb_prompt}. Our motivation is that given GPT-4 is highly capable of natural language understanding, a few-shot prompting style used in \cite{yu2023metamath} might suppress the diversity of the augmented questions. The seed dataset is constructed by the samples in the training set of MATH that do not contain Asymptote language in their question statements. We perform rejection sampling from GPT-3.5 on both the seed dataset and generated questions using the prompt shown in Figure~\ref{fig:cot_prompt}, obtaining 66.5K question-response pairs. We use a temperature $T=1.0$ for both question bootstrapping and rejection sampling.

\textbf{Augmented Similar Problems. } With the same seed dataset, we ask GPT-4 to generate 3 problems (with a solution, for rejection sampling) for 1 seed problem each time, using the prompt in Figure~\ref{fig:sim_prompt}. This is different from the practice in \cite{liu2023tinygsm}, where they ask GPT-3.5 to generate 10 similar questions given 1 seed problem since we find that GPT tends to generate several almost the same problems regardless of the given seed problem when asked to generate up to 10 new problems. We use the stronger GPT-4 instead of GPT-3.5 considering rejection sampling needs the answer to the problem better to be correct. To control the cost, our prompt emphasizes that the solution should be as brief as possible. The total number of the augmented similar problems and the question-response pairs rejection sampled from them is 38.2K. The rejection sampling prompt is the same one in Figure~\ref{fig:cot_prompt} as well. We use a temperature $T=1.0$ for both procedures.

\begin{figure*}[ht]
\begin{tcolorbox}
You will be provided with 1 math problem in newline-delimited json format. Please augment 5 diverse problems from the given problem.\\

The way you augment a problem can be:\\
- Rephrase the problem.\\
- Change the scenario without modifying specific quantities.\\
- Set 1 number in the problem to an unknown variable, put the answer in the problem and ask what is the value of the variable. Ensure the generated problem is reasonable. Otherwise, skip this method.\\
- Other approaches that can ensure the correctness of the answer you provide to the augmented problem.\\

Your response should only contain text in newline-delimited json format, keeping the same with the given problem. Please use two backslashes to represent one in the strings.

\end{tcolorbox}
\caption{The prompt we use to perform question bootstrapping for asking GPT-4.}
\label{fig:qb_prompt}
\end{figure*}

\begin{figure}[ht]
\begin{tcolorbox}

You will be presented a mathematical problem. You should solve the problem step-by-step carefully. Present the final answer in latex boxed format, e.g., $\boxed{63\pi}$.

\end{tcolorbox}
\caption{The prompt we use to do rejection sampling from GPTs.}
\label{fig:cot_prompt}
\end{figure}

\begin{figure}[ht]
\begin{tcolorbox}
You will be provided with 1 math problem in newline-delimited json format. Please generate 3 diverse new problems similar to the given problem. \\

Your response should only contain text in newline-delimited json format, keeping the same with the given problem. The solutions to the generated problems should be as brief as possible. Ensure there is only one box in the solution and the answer is completely the same with the content in the box. Please use two backslashes to represent one in the strings.
\end{tcolorbox}
\caption{The prompt we use to generate questions similar to the seed problems for asking GPT-4.}
\label{fig:sim_prompt}
\end{figure}

\textbf{Iterative Question Composing. }
We perform Iterative Question Composing for 4 iterations as described in Section~\ref{subsec:iqc}. Specifically, we use GPT-4 for question composing model $\pi_q$ with a $T=0.7$ temperature and GPT-3.5 for rejection sampling model $\pi_r$ with a $T=1.0$ temperature. The question composing prompts and rejection sampling prompt are shown in Figure~\ref{fig:qc_prompt} and Figure~\ref{fig:cot_prompt}, respectively. The text templater $x(\cdot, \cdot)$ we use is a program that transforms each question-response pair into JSON text format, with fields `problem' and `solution'. The seed dataset is also the samples in the training set of MATH that do not contain Asymptote code in their question statements. The resulting dataset has 55.1K samples in total.\footnote{A part of the samples are generated by performing IQC for 2 iterations using a legacy version of prompts.} We provide an example of the generated questions in different iterations corresponding to the same seed problem in Figure~\ref{fig:iqc_example}. We note that although some of the questions are not rigorously a sub-problem or sub-step of the corresponding problem in the previous iteration as required in our prompt, they are still valid questions that can increase the diversity of the dataset. We have checked the correctness of 100 randomly selected QA pairs generated by IQC and find that 85\% of them are correct.

\textbf{Mathematics Stack Exchange.} We observe that in the OpenWebMath~\cite{openwebmath} dataset, the data from Mathematics Stack Exchange shows high quality and is most related to competition-level math. Motivated by this, we extract the data collected from Mathematics Stack Exchange in RedPajama~\cite{together2023redpajama} and pre-process it into question-response pairs. For each Mathematics Stack Exchange page, we only retain the answer ranked first by RedPajama. Then we filter out the answer that does not contain a formula environment symbol `\$'. This results in a dataset with 1203.6K question-response pairs.

\begin{table}[t]
\caption{The composition of MMIQC.}
\label{tab:mmiqc}
\vskip 0.05in
\begin{center}
\begin{small}
\begin{sc}
\begin{tabular}{lcccr}
\toprule
Data              & \#~Samples & \#Repetitions & Ratio \\
\midrule
MetaMathQA & 203.7K & 3 & 26.6\%\\
AnsAug \& QB       & 66.5K & 3 &  8.7\%\\
AugSimilar         & 38.2K & 3 &  5.0\%\\
\textbf{IQC}                & \textbf{55.1K} & 3 &  \textbf{7.2\%} \\
MathStEx & 1203.6K & 1 & 52.5\%\\
\bottomrule
\end{tabular}
\end{sc}
\end{small}
\end{center}
\vskip -0.1in
\end{table}

Table~\ref{tab:mmiqc} shows the make-up of MMIQC. When fine-tuning the models MMIQC contains 3 repetitions of the subsets mentioned above, except for the Mathematics Stack Exchange part. We shuffle the order of samples after combining the subsets.

\section{Experiments}
\label{sec:exp}

\subsection{Fine-tuning Setup}
\label{subsec:ft-setup}
Our fine-tuning strategy mainly follows the practice of \cite{alpaca}, except that we use a different prompt template to transform the question-response pairs. For a sample from Mathematics Stack Exchange, the corresponding prompt fed into the model during training is a simple concatenation of the question and response with two new-line symbols. For a sample from other subsets, we additionally add a prefix `Please solve the following problem and put your answer at the end with ``The answer is: ''.' to the question-response concatenation.

We use the HuggingFace transformers library~\cite{wolf2019huggingface} for our fine-tuning experiments.
\begin{table}[t]
\caption{Ablation study on the optimal learning rate. We fine-tune Mistral-7B on MMIQC with different maximal learning rate values and evaluate the fine-tuned models on MATH to decide the best candidate.}
\label{tab:lr_select}
\vskip 0.05in
\begin{center}
\begin{small}
\begin{sc}
\begin{tabular}{lccccccr}
\toprule
LR & 1e-6 & 5e-6 & \textbf{1e-5} & 2e-5 & 5e-5 & 1e-4 \\
\midrule
MATH(\%) & 32.3 & 35.1 & \textbf{36.0} & 35.4 & 31.5 & 27.1\\
\bottomrule
\end{tabular}
\end{sc}
\end{small}
\end{center}
\vskip -0.3in
\end{table}

We fine-tune all models on MMIQC for 1 epoch, using a $3\%$ warm-up ratio linear learning rate schedule. For the choice of maximum learning rate, we do a simple hyper-parameter selection experiment shown in Table~\ref{tab:lr_select} and determine it to be 1e-5.
We use the BFloat16 numerical format during training. Employing the DeepSpeed Zero-3 Stage~\cite{rajbhandari2020zero}, we fine-tune 7B models on one node of 8xA800 GPUs with micro batch-size at 8, and gradient accumulation at 4, 34B models on 2 nodes with micro batch-size at 4 and gradient accumulation at 4 and $\sim$70B models on 4 nodes with micro batch-size at 4 and gradient accumulation at 2, 
maintaining an effective batch size of 256. It takes around 14 hours, 61 hours and 90 hours to fine-tune 7B, 34B and $\sim$70B models under the setups stated above, respectively. 

\begin{table*}[t]
\vspace{-10mm}
\caption{A comparative analysis of the accuracies achieved by various models on the MATH benchmark. The models marked with an asterisk($*$) are fine-tuned and evaluated by us.
Other results, unless otherwise cited, are derived from \cite{wang2023math}. This comparison highlights the significant improvements our fine-tuned models demonstrate over existing solutions in mathematical problem-solving accuracy.}
\label{tab:math-acc}
\vskip -0.15in
\begin{center}
\begin{small}
\begin{sc}
\resizebox{1.0\textwidth}{!}{%
\begin{tabular}{lcccr}
\toprule
Model      & FT-Dataset & Tool Usage? & Eval Method  &  MATH(\%) \\
\midrule
proprietary models \\
\midrule
\quad Minerva-540B~\citep{uesato2022solving} & Arxiv+Web & No & maj1@64 &  50.3 \\
\quad GPT-4 {\tiny(2023-0314)}~\citep{bubeck2023sparks} & - & No & pass@1 & 42.5 \\
\quad Gemini-Ultra~\citep{team2023gemini} & - & No & pass@1 & 53.2\\
\midrule
$\sim$7B models \\
\midrule
\quad Llama-2-7B~\citep{Touvron2023Llama2O} & - & No & pass@1 & 2.5 \\
\quad Qwen-7B~\citep{bai2023qwen} & - & No & pass@1 & 11.6 \\
\quad Llemma-7B~\citep{azerbayev2023llemma} & Proof-Pile-2 & No & pass@1 & 18.0 \\
\quad MetaMath-7B~\citep{yu2023metamath} & MetaMathQA & No & pass@1 &  19.8 \\
\quad Mistral-7B-MetaMathQA~\citep{yu2023metamath}  & MetaMathQA & No  & pass@1 & \underline{28.2}  \\
\quad \textbf{Mistral-7B-MMIQC*}       & \textbf{MMIQC} & No & pass@1    &  \textbf{36.0} \\
\quad MAmmoTH-Coder-7B~\citep{yue2023mammoth} & MathInstruct & Code & pass@1 & 35.2 \\
\quad ToRA-Code-7B~\citep{gou2023tora}  & ToRA-Corpus & Code & pass@1 &  44.6 \\
\midrule
$\sim$34B models \\
\midrule
\quad CodeLlamma-34B & - & Code  & pass@1 & 25.0 \\
\quad Llemma-34B-MetaMathQA & MetaMathQA & No  & pass@1 & \underline{34.8} \\
\quad \textbf{Llemma-34B-MMIQC*} & \textbf{MMIQC} & No & pass@1 & \textbf{38.6} \\
\quad Llemma-34B-MetaMathQA & MetaMathQA & Math-Shepherd  & maj+verify1@256 &  47.3 \\
\quad ToRA-Code-34B~\citep{gou2023tora}  & ToRA-Corpus & Code & pass@1 &  50.8 \\
\midrule
$\sim$70B models \\
\midrule
\quad Llama-2-70B~\citep{Touvron2023Llama2O} & - & No & pass@1 & 13.5 \\
\quad DeepSeek-67B~\citep{bi2024deepseek} & - & No & Pass@1 & 18.7 \\
\quad Deepseek-67B-MetaMathQA  & MetaMathQA &  No  & pass@1 &  \underline{36.8} \\
\quad \textbf{Deepseek-67B-MMIQC*}  & \textbf{MMIQC} &  No  & pass@1 &  \textbf{41.0} \\
\quad Deepseek-67B-MetaMathQA  & MetaMathQA &  No  & maj1@256 &  45.4 \\
\quad Deepseek-67B-MetaMathQA  & MetaMathQA &  Math-Shepherd  & maj+verify1@256 &  48.1 \\
\quad Qwen-72B~\citep{bai2023qwen}  & - &  No  & pass@1 &  35.2 \\
\quad Qwen-72B-MetaMathQA*  & MetaMathQA &  No  & pass@1 &  \underline{41.7} \\
\quad \textbf{Qwen-72B-MMIQC*}  & \textbf{MMIQC} &  No  & pass@1 &  \textbf{45.0} \\
\bottomrule
\end{tabular}
}
\end{sc}
\end{small}
\end{center}
\vskip -0.1in
\end{table*}

\subsection{Model Evaluation}

For a fair comparison, we first evaluate the fine-tuned models on MATH~\cite{hendrycks2021measuring}, a competition-level math word problems benchmark with 5000 test problems in a \textbf{zero-shot} setting. We prompt all our fine-tuned models with the test question with the prefix `Please solve the following problem and put your answer at the end with ``The answer is: ''.', and extract the answer from the output using a modified version of the answer extractor provided in \cite{lewkowycz2022solving}. We use a series of rules to infer whether the extracted answer is the same as the ground-truth answer, including a comparison using SymPy~\cite{sympy}. The complete results of our evaluation on MATH and a comparison with existing models are shown in Table~\ref{tab:math-acc}.

For the evaluation on 2023 Hungarian national high school finals in mathematics, we use the few-shot prompt used in \cite{Keiran2024hungarian}. We manually assess the grades for every model according to the examiner instructions. The results shown in Figure~\ref{fig:hungarian} are the grades under a full mark of 117.

\subsection{Ablation Study on Subsets of MMIQC}
\begin{table}[t]
\caption{How different subsets of MMIQC affect the accuracy of the finetuned model on MATH. }
\label{tab:ablation}
\begin{center}
\begin{small}
\begin{sc}
\begin{tabular}{lcccr}
\toprule
Data              & \#~Samples &  MATH(\%) \\
\midrule
MetaMathQA & 395K & \underline{28.2} \\
\midrule
MetaMathQA (subset) & 203.7K & 26.4 ({-1.8}) \\
+ AnsAug \& QB      & +66.5K & 30.1 ({+1.9}) \\
+ AugSimilar       & +38.2K & 31.5 ({+3.3}) \\
+ \textbf{IQC} iter~\#1        & +21.8K & 33.0 ({+4.8})  \\
+ \textbf{IQC} iter~\#2        & +16.0K & 33.7 ({+5.5})  \\
+ \textbf{IQC} iter~\#3 \& \#4     & +17.3K & 34.4 ({+6.2}) \\
+ MathStackExchange  & +1203.6K & \textbf{36.0} ({\textbf{{+7.8}}}) \\
\bottomrule
\end{tabular}
\end{sc}
\end{small}
\end{center}
\end{table}

In order to understand the ratio of contribution to the improvement revealed in Table~\ref{tab:math-acc} of different subsets of MMIQC, we fine-tune Mistral-7B with a series of training sets constructed by gradually adding the subsets. When MathStackExchange is not added, we fine-tune for 3 epochs. When MathStackExchange is added to the training dataset, we mix 3 repetitions of other data with 1 repetition of the MathStackExchange, and fine-tune for only 1 epoch. It can be seen from Table~\ref{tab:ablation} that 
\begin{itemize}
    \item Although our filtered subset of MetaMathQA is only half the size of the original dataset (which has 395K samples, more than the total number of samples of our synthetic data), the performance drop is only 1.8\%. This shows that the $k=20$ strategy in \cite{yu2023metamath} results in some redundancy.
    \item Our Answer Augmentation \& Question Boosting data help the fine-tuned model beat Mistral-7B-MetaMathQA, verifying our hypothesis that directly asking GPT to perform question bootstrapping is more efficient than providing few-shot examples to them.
    \item Our IQC method leads to a significant 3.1\% improvement from a high accuracy of $31.5\%$ with only 55.1K samples, showing its efficiency. Moreover, the later iterations of IQC also account for a certain ratio of improvement, proving that IQC is a method that can continuously generate new data that can help increase the diversity when added to the data generated in previous iterations.
\end{itemize}

\subsection{Contamination Test}
We check the $n$-gram matches for MMIQC to ensure that the improvement is not a result of direct memorization. We use the script provided by \cite{azerbayev2023llemma} to check the $n$-gram matches between the synthetic part of the MMIQC and MATH test set. It turns out that for a 30-gram match check, there are 44 hits of match between the `solution' field of MATH test set and the `output' field of MMIQC, far fewer than the 168 hits between that of MATH test set and MATH training set. Moreover, we manually check these 44 hits and find that 43 among them belong to the case where intermediate steps of the solutions to similar but different questions collide, with the only exception being the question `A regular polygon has interior angles of 144 degrees. How many sides does the polygon have?'. This almost rules out the possibility that fine-tuned models get memorization of solutions to the problems in the test set, indicating a very low risk of data contamination for MMIQC.

\section{Conclusion}

In this work, we introduce a novel data augmentation method for math word problem datasets called IQC (Iterative Question Composing) and use it in the construction of our MMIQC dataset. Our evaluation results show that the models fine-tuned on MMIQC achieve new SOTAs on the MATH benchmark. The improvements of our models benefit from the diverse data sources of MMIQC and the effectiveness of IQC. 

For future directions, we are interested in how to equip open-source models with the ability to compose questions, in order to perform IQC in a self-evolution style, similar to that in \cite{huang2022large}. Besides, how to integrate the verification systems~\citep{wang2023math, liu2023tinygsm} that are originally used to improve the accuracy during inference time into the procedure of IQC, is also an attractive topic.

\section*{Acknowledgements}
We thank Yang Yuan, Kaiyue Wen, Xingyu Dang, and Jingqin Yang for their helpful discussions.

\bibliography{reference}

\end{document}